%% file: main.tex
\documentclass[sigconf,,authorversion,nonacm]{acmart}
\usepackage{subfigure}
\usepackage{multirow}
\AtBeginDocument{%
  \providecommand\BibTeX{{%
    \normalfont B\kern-0.5em{\scshape i\kern-0.25em b}\kern-0.8em\TeX}}}

\begin{document}
\newcommand{\huanle}[1]{{\color{blue}\textsf{#1}}}

\title{Early Mobility Recognition for Intensive Care Unit Patients Using Accelerometers}

\author{Rex Liu}
\email{rexliu@ucdavis.edu}
\orcid{0000-0001-9642-8156}
\affiliation{%
  \institution{Department of Computer Science\\
  University of California, Davis}
  \streetaddress{1 Shields Ave}
  \city{Davis}
  \state{California}
  \country{USA}
  \postcode{95616}
}

\author{Sarina A Fazio}
\email{safazio@ucdavis.edu}
\affiliation{%
  \institution{
  Department of Internal
  Medicine\\Division of Pulmonary, Critical Care, and Sleep Medicine\\
  University of California, Davis}
  \streetaddress{2315 Stockton Blvd., North Addition 3030-7}
  \city{Davis}
  \state{California}
  \country{USA}
  \postcode{95817}
}

\author{Huanle Zhang}
\email{dtczhang@ucdavis.edu}
\affiliation{%
  \institution{Department of Computer Science\\
  University of California, Davis}
  \streetaddress{1 Shields Ave}
  \city{Sacramento}
  \state{California}
  \country{USA}
  \postcode{95616}
}

\author{Albara Ah Ramli}
\email{arramli@ucdavis.edu}
\affiliation{%
  \institution{Department of Computer Science\\
  University of California, Davis}
  \streetaddress{1 Shields Ave}
  \city{Davis}
  \state{California}
  \country{USA}
  \postcode{95616}
}

\author{Xin Liu}
\email{xinliu@ucdavis.edu}
\affiliation{%
  \institution{Department of Computer Science\\ University of California, Davis}
  \streetaddress{1 Shields Ave}
  \city{Davis}
  \state{California}
  \country{USA}
  \postcode{95616}
}

\author{Jason Yeates Adams}
\email{jyadams@ucdavis.edu}
\affiliation{%
  \institution{
  Department of Internal Medicine \\
  Division of Pulmonary, Critical Care, and Sleep Medicine \\ University of California, Davis}
  \streetaddress{2315 Stockton Blvd., North Addition 3030-7}
  \city{Davis}
  \state{California}
  \country{USA}
  \postcode{95817}
}


\begin{abstract}
With the development of the Internet of Things(IoT) and Artificial Intelligence(AI) technologies, human activity recognition has enabled various applications, such as smart homes and assisted living. In this paper, we target a new healthcare application of human activity recognition, early mobility recognition for Intensive Care Unit(ICU) patients. Early mobility is essential for ICU patients who suffer from long-time immobilization. Our system includes accelerometer-based data collection from ICU patients and an AI model to recognize patients' early mobility. To improve the model accuracy and stability, we identify features that are insensitive to sensor orientations and propose a segment voting process that leverages a majority voting strategy to recognize each segment's activity. Our results show that our system improves model accuracy from 77.78\% to 81.86\% and reduces the model instability (standard deviation) from 16.69\% to 6.92\%, compared to the same AI model without our feature engineering and segment voting process.
\end{abstract}




\maketitle

\input{introduction}

\input{related_work}

\input{dataset}

\input{design}

\input{result}

\input{discussion}

\input{conclusion}

\bibliographystyle{ACM-Reference-Format}
\bibliography{references}

\end{document}

%% file: introduction.tex
\section{introduction}
Due to long periods of inactivity and immobilization, Intensive Care Unit(ICU) patients become weak when recovering from major illnesses~\cite{b01}. Early Mobility(EM) is an effective and safe intervention to improve functional outcomes of ICU patients, which decreases ventilator days and ICU lengths of stay~\cite{b03}. Therefore, it is of great interest for doctors to identify ICU patients' early mobilization. When doctors can accurately recognize the EM activities of the ICU patients, they can prescribe an optimal personalized dose of mobility to the ICU patients. However, the advancement of research on EM is limited due to the lack of accurate and effective systems to recognize patients' EM activities in the ICU.

EM recognition is a sub-topic of Human Activity Recognition(HAR), a fast-moving area because of the recent advancement of  Internet of Things(IoT) and Artificial Intelligence(AI) technologies. Specifically, IoT technology enables convenient data acquisition and edge computing capacity, while AI technology provides accurate and efficient machine learning algorithms. Consequently, HAR systems have been applied in various fields such as fitness, smart homes, and assisted living~\cite{b04}. HAR healthcare applications facilitate disease detection and provide proactive assistance for both patients and doctors, etc. For example, \citeauthor{b06} built a HAR system to analyze the symptoms of Parkinson patients~\cite{b06};
 \citeauthor{b05} built a proactive assistance system for senior citizens by identifying their abnormal activities~\cite{b05}. 

Based on the type of data being collected, a HAR system can be classified into vision-based \cite{b30}, and motion sensor-based \cite{b14}. A vision-based HAR system usually deploys cameras at fixed locations of interest and applies computer vision techniques to recognize human activities. However, such systems have severe privacy issues, especially in the ICU. In comparison, a sensor-based HAR system leverages lightweight sensors that patients can carry. Different types of sensors can capture different features of activities, e.g., movements by accelerometer and gyroscope \cite{b33}, environment by temperature and humidity sensors \cite{b32}, and physiological signals (e.g., heart rate) by electrocardiogram sensors \cite{b31}. We leverage accelerometers to recognize ICU patients' EM activities because accelerometers are lightweight, energy-efficient, and widely available (e.g., in smartphones).

In our system, each patient wears two accelerometer sensors: one sensor is on the chest, and the other sensor is on the thigh. Patients perform routine activities in the ICU room while the sensor data is continuously collected. Our system automatically recognizes patients' EM activities based on the sensor data. Since our system targets applications in ICU, it needs to have both high \textbf{accuracy} and \textbf{stability}. High accuracy means that the system achieves satisfactory average recognition precision of all patients, while high stability emphasizes that the recognition accuracy for different patients is approximately the same. 

There are three main challenges in realizing our system. 
(1) \emph{Distorted Activities}.  EM activities are different from traditional human activities in the HAR system. Traditional human activities usually consist of regular activities with periodic patterns, such as running, walking, and swimming~\cite{b34}. In comparison, ICU patients' EM activities are distorted, take much longer than regular activities, and vary among patients. Also, they may be affected by medical equipment, and physical therapy,  such as Range Of Motion(ROM) exercises with the therapist's assistance, which is an EM activity that helps patients regain joint or muscle strength.
(2) \emph{Label Noise}. A medical expert annotates the activity of each minute's sensor data. However, for ICU patients with different health conditions, the time needed to accomplish an activity is significantly different. Furthermore, intermittent behaviors during the activities are observed. For example, one patient could finish an EM activity once in a minute while another patient could accomplish the same EM activity twice in a minute or accomplish this same EM activity and other non-EM activity in a minute. Hence, the label for every minute's sensor data is coarse, leading to the label noise. Label noise is an important issue for classification tasks, which decreases the accuracy and the stability of AI models \cite{b08}. 
(3) \emph{Sensor Orientation.}
Although sensors are placed at roughly the exact locations (thigh and chest) on our ICU patients, their orientations are different. As a result, the 3-axis sensor readings (X, Y, Z-axis) have different physical meanings for different patients. Therefore, without careful feature extraction and selection, the AI model generalizes poorly to different patients. 

To the best of our knowledge, our work is one of the first HAR systems for recognizing ICU patients' EM activities using wearable sensors. In realizing our system, we make the following contributions. (1) A system for EM recognition for ICU patients. Our system adopts two accelerometer sensors in different positions to capture different movement features of EM activities. Through a combination of sensors on the chest and on the thigh, the system collects more information to identify distorted movements.  (2) We propose a segment voting process to handle the label noise problem. Specifically, each one-minute sensor data is divided into multiple fixed half-overlapped sliding segments (time windows). We train our AI model using the segments. To predict the activity of each one-minute sensor data,  we apply our trained model to each segment. The final prediction result for the one-minute data is the activity that has the majority vote among the predictions of all segments. 
(3) To tackle the sensor orientation problem, we identify and extract features that are not sensitive to sensor orientations. Our features improve both the accuracy and the stability of AI models compared to the model trained on commonly-used features. 

We evaluate the accuracy and the stability of our system for classifying two categories of lying activities for ICU patients. To objectively evaluate our system for new patients, we adopt leave-one-patient-out cross-validation. The experimental results show that our system increases the classification accuracy from 77.78\% to 81.86\% and reduces the model instability (standard deviation) from 16.69\% to 6.92\%, compared to the model without our feature engineering and segment voting process.

%% file: related_work.tex
\section{Related Work}

This section presents related work on machine learning algorithms for HAR, HAR applications in the healthcare domain, and the community's endeavor to assist ICU patients. 

\vspace{0.1in}
\noindent
\emph{Machine Learning Algorithms for HAR}.  Different machine learning algorithms (classical machine learning and deep learning) are adopted for recognizing activities. For example, classical machine learning algorithms such as Decision Tree \cite{b35} and Logistic Regression \cite{b15} are applied to the extracted clinical features. Deep learning algorithms such as convolution neural networks are recently leveraged, which achieve higher accuracy \cite{b17}. We focus on classical machine learning algorithms because 1) our collected dataset is not enough for deep learning (only 586 data samples), and 2) doctors prefer feature-based algorithms in order to interpret the results.

\vspace{0.1in}
\noindent
\emph{HAR for Healthcare Applications}. 
Different medical applications face different challenges and considerations, so various HAR systems are designed to target different applications. For example, \citeauthor{b27} proposed a Support Vector Machine (SVM)-based approach to evaluate the severity of symptoms for patients with Parkinson’s disease based on wearable sensor data~\cite{b27}; \citeauthor{b28} proposed a neural network classifier to assist the recovering of patients by recognizing squats (a common rehabilitation exercise)~\cite{b28}; \citeauthor{b29} targeted aging populations in the home to assist living such as abnormal activities detection and security assurance~\cite{b29}. As far as we know, our system is one of the first works on HAR for ICU patients using wearable sensors. 

\vspace{0.1in}
\noindent
\emph{HAR for ICU Patients}. 
There are several works that leverage cameras to assist ICU Patients. For example,  \citeauthor{b22} developed computer vision algorithms to detect patient mobilization activities in the ICU, which achieves a mean area under the curve of 0.938 for identifying four types of activities~\cite{b22};  \citeauthor{b23} used depth cameras to mitigate the severe privacy issue of RGB cameras, and identify two types of EM activities \cite{b23}. In comparison, we use accelerometer sensors to avoid intrusive cameras in ICU environments. In addition, our system is less costly than the camera-based systems.  

%% file: dataset.tex
\section{Data Collection}

\begin{figure}[!t]
\centerline{
\subfigure[Sensor position]{%
\begin{minipage}[b]{1.6in}
\centering
\label{fig:device_setting}
\includegraphics[height=1.6in]{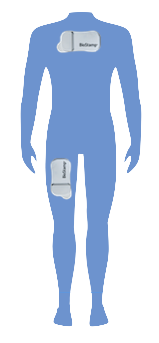}
\end{minipage}
}%
\subfigure[Accelerometer sensor]{%
\begin{minipage}[b]{1.6in}
\centering
 \label{fig:sensor-fig}
\includegraphics[width=1.2in]{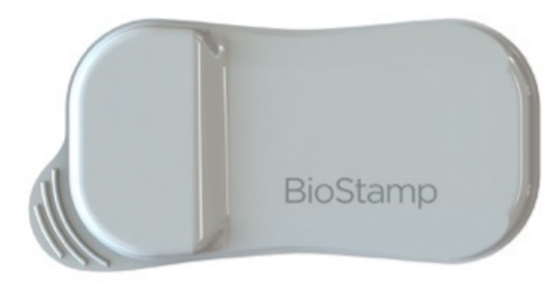}
\end{minipage}
}%
}
\caption{
The sensor setup in our EM recognition system. (a) Two accelerometer sensors are worn by ICU patients. One sensor is on the chest, and the other sensor is on the thigh. (b) The clinical-grade 3-axis accelerometer sensor that is used in our system. 
}
\label{fig:device_setup}
\end{figure}

EM activities data were collected in individual patient rooms in a Medical ICU (MICU) room of an academic medical center in California. Seventeen adult MICU patients who are eligible for early mobility interventions between 2016-2017 were recruited for data collection.
As Figure \ref{fig:device_setup} illustrates, two 30Hz clinical-grade 3-axis accelerometer sensors are placed on each participant: one sensor is attached to the chest, and the other sensor is attached to the thigh. 
Generally, continuous accelerometer data were collected over a period of 4 hours and up to 48 hours to ensure representative sampling of day and night mobility interventions and patient activities. Concurrently, all ICU activities were recorded using a camera system without audio. After the conclusion of the monitoring period, an ICU clinician reviewed all videos and annotated each minute's sensor data EM activity based on the corresponding video. However, due to device and timing reasons, video data or sensor data were missing for nine patients. Only eight patients with 586 minutes of sensor data for lying EM activities are annotated.

We classify two categories of lying activities as suggested by the American Association of Critical-Care Nurses (AACN) and the Society of Critical Care Medicine (SCCM), who define the quantification approach of ICU early mobility \cite{b07}. Specifically, we focus on lying without EM activities and lying with EM activities.  
The lying with EM activities includes four EM activities when patients lie on the bed; that is, repositioning,  Range Of Motion(ROM) exercises, percussion therapy, and suction, cough, and oral care.  Repositioning involves lying flat in bed and moving laterally from left to right side or vice versa; ROM exercises help recover patients' muscles and joints; percussion therapy assists patients with removing respiratory fluid tapped on the back of the patients' chest for 8-10 minutes by clinicians; suction, cough and oral care are the daily activities in bed for ICU patients.

%% file: design.tex
\section{System Architecture \& EM Classification}

In this section, we first provide an overview of our system. Then, we explain our segment voting process and the features used in our system. Last, we introduce our machine learning model for classifying lying EM activities. 

\begin{figure}[!t]
\centering
\centerline{
\includegraphics[width=3.0in]{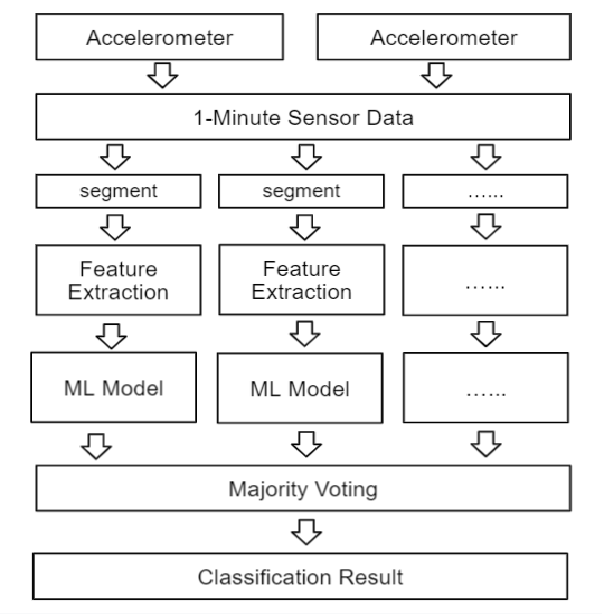}}
\caption{
Data flow of our EM recognition system.
}
\label{fig:framework}
\end{figure}

\subsection{System Overview}

Figure \ref{fig:framework} shows the data flow of our EM recognition system. Our system takes 1-minute sensor data from the two accelerometers as input. The data from these two sensors are synchronized and integrated. Afterward, each 1-minute sensor data is divided into multiple half-overlapped sliding segments (time windows). Then, we extract time-domain and frequency-domain features that are insensitive to sensor orientations. To mitigate the noisy label problem of EM activities, we leverage the segment voting process, which predicts the activity of the 1-minute input as the activity that has the majority vote of segments' prediction results.  

\subsection{Extracted Features from Sensor Data}
Both time-domain features and frequency-domain features are widely used in HAR systems~\cite{b09, b10, b14}. Time-domain features reflect the trend of sensor signal changing with time, which can help the EM recognition system to capture the changes in activities \cite{b11}. In contrast, frequency-domain features capture dynamic motion in an activity, which can improve the EM recognition system to get the movement trend in an activity~\cite{b13}. We use Fast Fourier Transform (FFT) to convert the sensor data from the time domain to the frequency domain to extract frequency-domain features. We extract the low-frequency components whose frequencies are less than 0.3Hz related to gravity's influence and the high-frequency components with frequencies between 0.3Hz and 20Hz related to the dynamic motion. And we also extract the derivation of each high-frequency component as our third frequency-domain features.  Because different patients have different and unknown sensor orientations, we cannot rely on features that are sensitive to sensor orientations~\cite{b02}, such as features related to each axis.  Instead, we apply the magnitude function to each tri-axial signal: original tri-axial signals, low-frequency components, high-frequency components, and the derivation of each high-frequency component. Finally,  for each sensor position, we have one time-domain feature, namely the signal magnitude,  and three frequency-domain features, namely the magnitude of low-frequency components, high-frequency components, and the derivation of each high-frequency component ,which are less susceptible to sensor orientation. Furthermore, we apply eight metrics (i.e., mean, maximum, minimum, standard deviation, median, and entropy) to each time and frequency-domain feature, resulting in a total of 64 attributes for training machine learning models.

\subsection{Classification Model}

Different machine learning algorithms have various advantages and disadvantages. Classical machine learning algorithms are more straightforward to interpret based on the importance of features, which is especially helpful in the healthcare field. In comparison, deep learning algorithms do not require feature engineering and tend to have higher accuracy. However,  the amount of training data and the training overhead of a deep learning model can be enormous. Our collected EM activities dataset contains 538 minutes of sensor data, and thus the number of samples is only 538 because each sample is 1-minute long. Therefore, deep learning models are not applicable to our project because of the limited data samples. Instead, we use classical machine learning models such as Logistic Regression, Bagging Decision Tree, and Support Vector Machine. Our experiments show that Bagging Decision Tree (Bagging for short) works best for our needs. 
Bagging is an ensemble meta-algorithm that reduces the high variance of the decision tree, and thus improves the accuracy and stability of the decision tree algorithm. 

\subsection{Segment Voting Process}

The classification model takes each 1-minute data segment as input. Initially, the system accuracy is low because of the noisy label problem described earlier on in this paper. To improve the recognition accuracy and model stability, we propose a segment voting process.  In our segment voting process, 1-minute data is segmented into small pieces (segments) with a fixed half-overlap time window size. There exists a trade-off in selecting window size. When the time window size is short, each window is more likely to contain only one activity. However, the extracted features may be insufficient to distinguish different activities. On the other hand, when the time window size is long, each segment may contain multiple activities, and thus the recognition accuracy is reduced. Therefore, deciding an optimal time window size is crucial. Therefore, in our segment voting process, we explore different time-window sizes to find the optimal time window size for the EM activities dataset. After that, 1-minute sensor data is divided into data segments. Then we extract features from each segment and apply the classification model to each segment, which predicts the segment's activity. The activity of the 1-minute input is determined by selecting the activity that occurs most among the segments (i.e., majority voting).

%% file: result.tex
\section{Evaluation}
In this section, we extensively evaluate our system concerning the model accuracy and stability. In addition, we explore how the time window size, the sensor positions and numbers, the extracted features, and the number of patients in the training dataset affect the system performance.

\subsection{Implementation}
All experiments are implemented using scikit-learn and Keras package, running on a server with Nvidia Titan Xp GPU and Intel Xeon W-2155. To objectively evaluate our system for new patients, we adopt leave-one-patient-out cross-validation. The model \textbf{accuracy} is calculated by averaging the prediction accuracy for all patients, while the model \textbf{instability} is the standard deviation of the prediction accuracy among all patients. 

\begin{table}[!htbp]
\caption{
Classification accuracy and stability when different time window sizes are used for segments.}
\centering
\resizebox{8.2cm}{!}{
\begin{tabular}{ c | c | c c c c }
\hline
\multirow{2}{*}{} &
  Baseline &
  \multicolumn{4}{c }{\begin{tabular}[c]{@{}c@{}}Segment time window length
  \end{tabular}} \\ \cline{2-6} 
                                                         & 1-minute & 4 sec   & 10 sec           & 20 sec  & 30 sec \\ \hline
\begin{tabular}[c]{@{}c@{}}Accuracy \end{tabular} & 77.78\%  & 79.85\% & \textbf{81.86\%} & 77.85\% & 78.9\% \\ \hline
\multicolumn{1}{l|}{\begin{tabular}[c]{@{}l@{}}Instability \end{tabular}} &
  16.69\% &
  7.89\% &
  \textbf{6.92\%} &
  13.01\% &
  9.41\% \\ \hline
\end{tabular}
}
\label{time_windows}
\end{table}

\subsection{Segment Voting Process}

We propose a segment voting process to mitigate the noisy label problem. Our segment voting process divides each 1-minute sensor data into segments of smaller time windows. We evaluate the model performance when different time window sizes are used. We compare our performance with the baseline case that takes the whole 1-minute data as input.  

Table \ref{time_windows} tabulates our system performance and the baseline performance. We have the following observations: (1) Our segment voting process improves both the system accuracy and stability. This is because our segment voting process mitigates the effect of noisy labels. By adopting a 10-seconds time window, we increase the system accuracy from 77.78\% to 81.86\% and reduce the system instability from 16.69\% to 6.92\%.  (2) The time window length affects the system performance. When the time window size is too small, such as 4 seconds, the segment data is insufficient for the model to classify EM activities accurately. On the other hand, when the time window size is too large, such as 30 seconds, the accuracy also decreases because of overlapped activities in the segment. We find that a 10-seconds time window works best for our EM recognition. In the rest of the experiments, we set the time window size to 10 seconds. 

\begin{figure}[!t]
\centering
\centerline{
\includegraphics[width=3in]{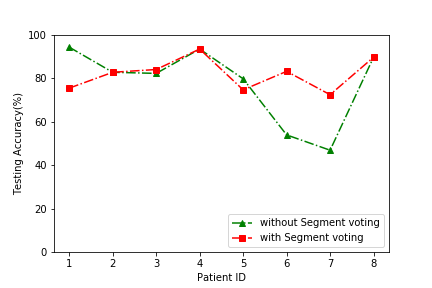}}
\caption{
Our segment voting process has better system stability. In other words, we achieve similar prediction accuracy for different patients. 
}
\label{fig:10s-comparision}
\end{figure}

Figure \ref{fig:10s-comparision} plots the prediction accuracy for each patient with our segment voting process and compares it with the baseline.  It clearly shows that our segment voting process results in better system stability than the baseline. Specifically, the baseline method has low prediction accuracy for patient ID 6 and 7, while our segment voting process still maintains good accuracy for both patients. System stability is essential for healthcare applications, which require consistent diagnosis results for patients.

\begin{table}[!htbp]
\caption{
The system performance for different sensor positions and sensor numbers.}
\centering
\resizebox{7cm}{!}{
\begin{tabular}{c|c c c}
\hline
Sensor &
  \begin{tabular}[c]{@{}c@{}}Chest only \end{tabular} &
  \begin{tabular}[c]{@{}c@{}}Thigh only \end{tabular} &
  \begin{tabular}[c]{@{}c@{}}Both \end{tabular} \\ \hline
Accuracy &
  70.28\% &
  80.29\% &
  \textbf{81.86\%} \\ \hline
Instability &
  16.98\% &
  9.51\% &
  \textbf{6.92\%} \\ \hline
\end{tabular}
}
\label{sensor_position}
\end{table}

\subsection{Extracted Features}

\begin{figure}[!t]
\centering
\centerline{
\includegraphics[width=3in]{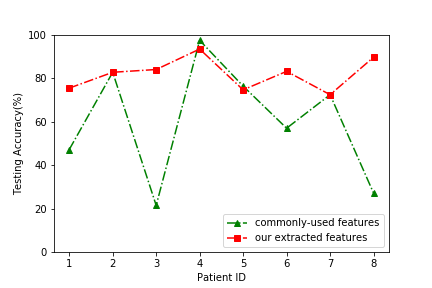}}
\caption{
We extract features that are insensitive to sensor orientation, which improves accuracy and stability more than using commonly-used features.
}
\label{fig:mag-feature}
\end{figure}

We investigate whether our extracted features perform better than commonly-used features. Precisely, we extract features that are not sensitive to sensor orientations, while existing systems often include more features such as features for each axis' sensor data. 

Figure \ref{fig:mag-feature} shows the performance of our extracted features which are only related to the magnitude of each tri-axial signal and commonly-used features, which contains the features of each axis. As we can see, our system has higher accuracy (especially for patients 1, 3, 6, and 8) and stability than the system with commonly-used features. Although the machine learning models can identify and extract the critical features, the limited dataset of ICU patients still requires careful feature engineering for the machine learning models to learn. 

\subsection{Sensor Positions and Numbers}

In our current system, we use two sensors (one on the chest and the other on the thigh) to recognize the EM activities of ICU patients. We explore whether our sensor setting is superior to using a single sensor. For comparison, we adopt the same machine learning model but use one sensor signal as input. 

Table \ref{sensor_position} tabulates the system accuracy and stability when using only one sensor on the chest, only one sensor on the thigh, and using two sensors. The results clearly show that our sensor setting (both sensors) achieves better accuracy and stability than the single sensor counterparts. Specifically, the thigh-only scenario is better than the chest-only scenario in recognizing lying activities. Meanwhile, by combining both sensor data, our scenario further increases 1.57\% accuracy and 2.59\% stability compared to the best case of a single sensor (i.e., thigh-only scenario).

\subsection{Number of Patients in Dataset}

\begin{table}[!htbp]
\caption{The system performance when different numbers of patients are used in the training and testing.
}
\centering
\resizebox{8cm}{!}{
\begin{tabular}{ c| c c c c }
\hline
\#Patients       & 5              & 6              & 7             & 8            \\ \hline
Accuracy           & 75.08\%        & 78.35\%        & 80.90\%       & 81.86\%      \\ \hline
Instability & 14.74\%   & 11.99\%   & 8.24\%   & 6.92\%  \\ \hline
\end{tabular}
}
\label{enumeration}
\end{table}

Our current system achieves 81.86\% accuracy and 6.92\% instability. Our system will improve if we have a larger dataset. To validate this hypothesis, we change the number of patients in our dataset and explore the system performance. 

Table \ref{enumeration} tabulates the system performance when the dataset size varies from 5 to 8. It clearly shows that by increasing the number of patients, our system performance improves. It also suggests that our dataset is not enough since the system performance has not yet reached the plateau. With a larger dataset, we expect that our system will improve further.

%% file: discussion.tex
\section{Discussion}

In this section, we discuss some limitations of our work and present future works. 

Our system adopts two accelerometer sensors and deploys them on the chest and thigh, which shows better accuracy and stability than the single sensor scenario. Clinicians may determine our sensor numbers and locations based on their experience. We believe that adding more sensors will improve the system performance at the expense of inconvenience for patients. The optimal number of sensors and sensor locations remains an open problem, especially for ICU patients with dramatically different activity patterns. 

Our system achieves 81.86\% accuracy, which may seem moderate considering the much higher accuracy reported by existing HAR papers. However, we target challenging activities of ICU patients, while existing works mostly focus on activities performed by healthy people who are consistent in conducting activities such as walking and running. In addition, from Table \ref{enumeration} we can see that our system has not fulfilled its full potential because the data sample size is still small. Our conjecture is that by collecting more data samples, our system's accuracy will increase further. 

In the future, we plan to consider more activities other than lying activities. As we have shown in this paper, recognizing EM activities of ICU patients involves many challenges, even for distinguishing only two categories of activities. Therefore, more research is required to support EM recognition for ICU patients using wearable sensors entirely.  

%% file: conclusion.tex
\section{Conclusion}

This paper presents the first work for EM recognition of ICU patients using wearable sensors. Due to ICU patients' irregular movement behaviors, EM recognition for ICU patients is more challenging than traditional human activity recognition (e.g., walking and running).  By systematically considering the sensor positions and numbers,  machine learning models,  extracted features, time windows, and segment voting process, our system achieves good accuracy (81.86\%) and stability (instability of 6.92\%) for recognizing two categories of lying activities of ICU patients. More research is required to support higher system performance and more activities of ICU patients.